\documentclass[letterpaper, 10 pt, conference]{ieeeconf}  

\IEEEoverridecommandlockouts                              

\usepackage{graphics} 
\usepackage{epsfig} 
\usepackage{mathptmx} 
\usepackage{times} 
\usepackage{amsmath} 
\usepackage{amssymb}  
\usepackage{subcaption}
\usepackage{kantlipsum} 
\usepackage{mwe}
\usepackage{multirow}

\DeclareMathOperator{\arctantwo}{arctan2}

\title{\LARGE \bf
BoxNet: A Deep Learning Method for 2D Bounding Box Estimation from Bird's-Eye View Point Cloud
}
\author{Ehsan Nezhadarya, Yang Liu and Bingbing Liu
\thanks{Ehsan Nezhadarya is with LG Canada {\tt\small e.n.arya@gmail.com} and he was with Huawei Technologies Canada when the work was done. Yang Liu and Bingbing Liu are with Noah's Ark lab, 2012
Laboratories, Huawei Technologies Canada. {\tt\small yang.liu6,
liu.bingbing@huawei.com}} %
}

\begin{document}

\maketitle
\thispagestyle{empty}
\pagestyle{empty}

\begin{abstract}

We present a learning-based method to estimate the object bounding box from its 2D bird's-eye view (BEV) LiDAR points. Our method, entitled BoxNet, exploits a simple deep neural network that can efficiently handle unordered points. The method takes as input the 2D coordinates of all the points and the output is a vector consisting of both the box pose (position and orientation in LiDAR coordinate system) and its size (width and length). In order to deal with the angle discontinuity problem, we propose to estimate the double-angle sinusoidal values rather than the angle itself. We also predict the center relative to the point cloud mean to boost the performance of estimating the location of the box. The proposed method does not rely on the ordering of points as in many existing approaches, and can accurately predict the actual size of the bounding box based on the prior information that is obtained from the training data. BoxNet is validated using the KITTI 3D object dataset, with significant improvement compared with the state-of-the-art non-learning based methods.

\end{abstract}

\section{INTRODUCTION}

Recent advances in autonomous driving have been marked by the extensive use of LiDAR. 
While some methods utilize a deep neural network to fuse both image and point cloud \cite{charles1,ku,chen} for improved performance, the others focus on LiDAR-only solutions for efficiency \cite{luo,yang,ren,zhou,charles2} with simpler network structures. On the other hand, LiDAR-only methods, both the end-to-end CNN-based solutions and the traditional ones that usually require sequential steps of segmentation, classification and tracking of the point cloud \cite{himmelsbach,wang,zermas}, will output a set of 3D object proposals represented with bounding boxes to incorporate the object geometric information such as size and orientation. Considering that the vertical dimension can be safely marginalized in some tasks including point cloud segmentation and object tracking in 2D, BEV is often used to reduce the size of data input in these applications. Our main focus in this paper is to estimate the object geometry from BEV point cloud, i.e. 2D bounding box that can be used in subsequent modules such as tracking and data association \cite{macLachlan}.

\begin{figure}[!t]
     \centering
      \includegraphics[scale=0.07]{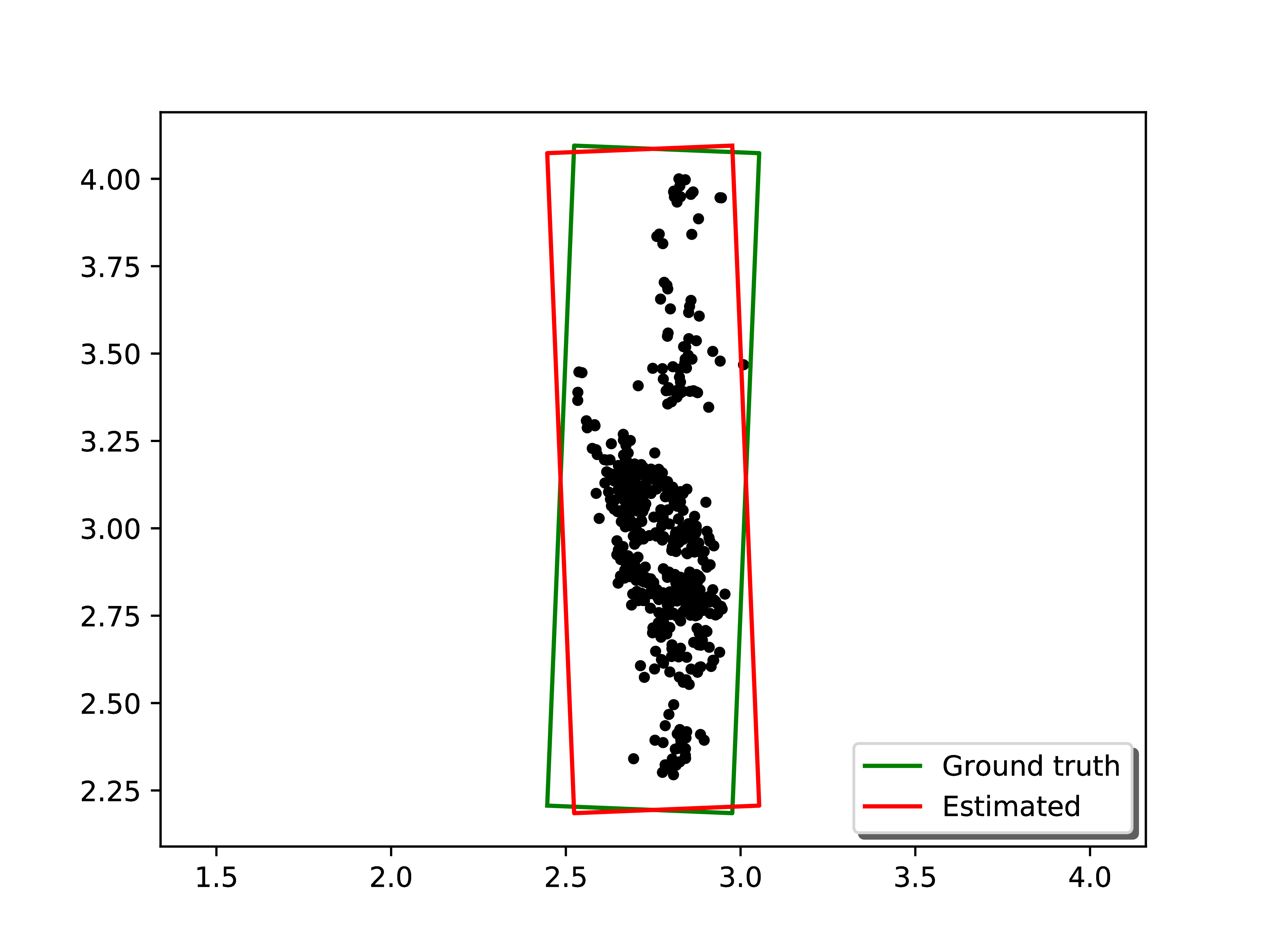}
      \caption{An example of angle discontinuity. The true difference between the estimated and ground truth bounding boxes is small. However, according to the definition domain of orientation, the ground truth is close to $\pi/2$ and the estimated one is close to $-\pi/2$, resulting in a large difference of almost $\pi$. The network is unable to directly learn the orientation with such an inconsistent angle description in training samples.}
      \label{discon}
\end{figure}

The proposed method uses a simple deep neural network which is designed to efficiently handle unordered points, and formulates bounding box estimation as a regression problem. The input to the network is a set of 2D unordered points, with the output being the object geometry, i.e. pose and size. By learning the correlation prior between points and geometry from training samples, accurate bounding box estimation can be achieved even when only partial points are observed by the sensor -- a common situation in autonomous driving. To resolve the discontinuity problem in defining the orientation (yaw angle) shown in Fig. \ref{discon}, we propose to use the double-angle sinusoidal values in the estimation. We also predict the offset from the object center to the point cloud mean for better estimation of the box location. Compared with the existing non-learning based methods, our proposed method does not rely on the ordering of points and can accurately predict the actual size of the box.

The rest of the paper is organized as following: Section \ref{literture} briefly discusses the background and literature, including 2D L-shape fitting and 3D bounding box estimation using deep learning. The proposed method is illustrated in Section \ref{method} with a demonstration of the network structure. We provide experimental validation on the KITTI 3D object dataset in Section \ref{experiment}, and conclude the paper in Section \ref{conclusion} with a short discussion.

\section{RELATED WORK}
\label{literture}

We omit the extensive literature review of 3D object detection methods, in which bounding box regression is considered as an essential component. Our proposed method only focuses on the 2D case and does not target an end-to-end solution. Therefore, it is in fact an intermediate step in the pipeline of 3D object detection. The goal is to provide accurate object bounding boxes for subsequent navigation and planning tasks. From this perspective, the problem defined in this paper is closely related to the L-shape fitting of 2D laser scanner data (or BEV LiDAR point cloud) in modeling a vehicle \cite{macLachlan,petrovskaya}.

In \cite{macLachlan}, a standard least-square method was used to fit both line and corner features, with a weighted procedure to discard outliers. Although details were not mentioned, this can be seen as the early strategy of fitting an L-shape. Petrovskaya {\it et al.} \cite{petrovskaya} used the technique of ``virtual scan" to reorganize the LiDAR point cloud for compact feature representation. Vehicle fitting was based on scan differencing with a probabilistic measurement model describing the occupancy of each grid along the virtual ray. A potential issue of virtual scan is that the information may be lost by quantizing the points. The authors in \cite{shen} solved the line fitting problem by formulating the points in a matrix and optimizing the parameters that correspond to the line coefficients. The time complexity is higher-order polynomial and the method requires an ordered set of points. In \cite{zhao}, a K-L transform was applied to the laser points to obtain the two axes that ideally correspond to the edges of the object. No estimation of width or length was mentioned in the method. Recently, three evaluation criteria were proposed in \cite{zhang} within an optimization framework to generate a bounding box that can embrace all the points. In \cite{kim}, the points were iteratively clustered to two orthogonal segments using their ordered information.

While these methods have different approaches to handle 2D points, they are either dependent on the ordering of points, or computationally expensive due to the clustering or optimization process. In addition, all the methods focus on the existing points that can be directly observed by the sensor, and therefore are unable to predict the actual size of the object in case that the object is only partially seen due to occlusion or sensor limitation. The major difference of our proposed method from the existing ones is that our method does not rely on the L-shape assumption for the 2D points. It is therefore more general and can be applied to various scenarios. On the other hand, the problem defined in this paper is different from the one addressed in many existing solutions to bounding box regression, in that we do not need to estimate the heading (direction) of the object because heading information cannot be reliably extracted from sparse 2D BEV LiDAR points. For this reason, we limit the bounding box orientation to the range $(-\frac{\pi}{2}, \frac{\pi}{2}]$.

\section{BoxNet for 2D Bounding Box Regression}
\label{method}

The proposed method focuses on learning the correlation prior between points and object geometry. It is partially inspired by the work in \cite{mousavian}, in which 3D bounding boxes are estimated from RGB images using deep neural network. Our method, however, addresses a similar problem in the case of 2D BEV LiDAR points. To enable fast learning and estimation, we exploit a simple network structure which is similar to that used in PointNet \cite{charles2}. We show how much a simple structure can give an accurate estimate of the 2D bounding box of an object from its 2D point cloud. To the best of our knowledge, this is the first time of investigating the possibility of using deep learning approaches in 2D bounding box estimation with BEV LiDAR points.

\subsection{BoxNet Architecture}
\begin{figure*}[t]
\begin{center}
\includegraphics[scale=0.45]{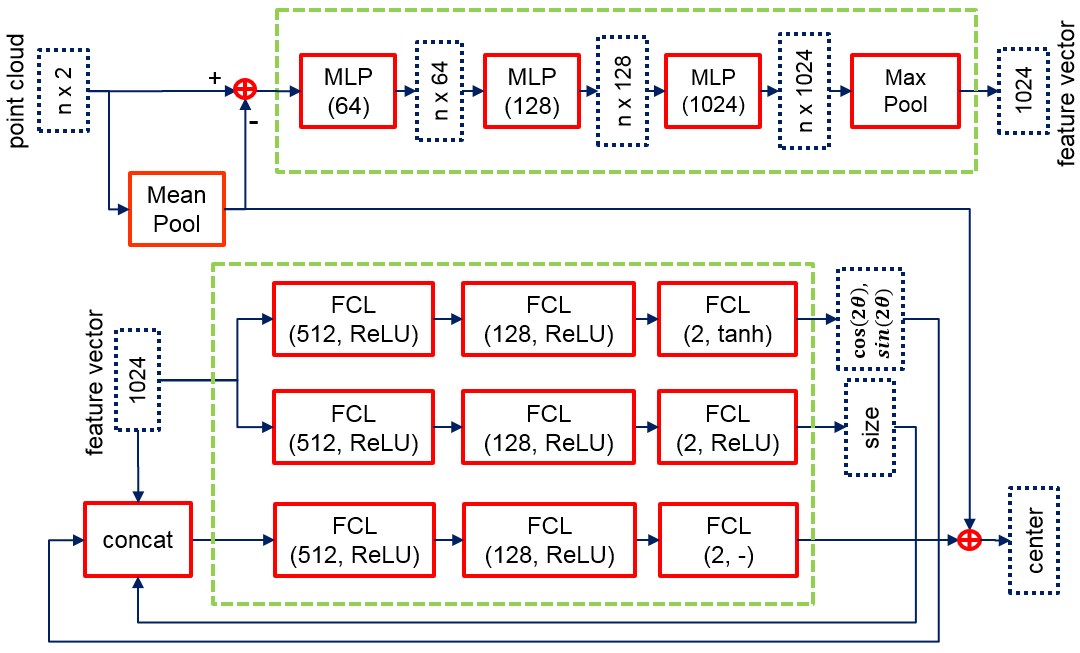}
\caption{\label{fig:boxnet} {\bf BoxNet architecture}. The top sub-network extracts a network a feature vector of size 1024 from a 2D point cloud of size $N$. The structure of fully connected layers used to estimate the bounding box angle, size and center from the feature vector is shown in the bottom sub-network.}
\end{center}
\end{figure*}

Architecture of the proposed BoxNet deep network is shown in Fig.~\ref{fig:boxnet}. The network input is a point cloud of $N$ points, in which each point is represented as the 2D BEV coordinate $(x,y)$. The network output is the vector $[\tilde{c}_x, \tilde{c}_y, w, l, \cos(2\theta), \sin(2\theta)]^T$, where $(\tilde{c}_x$, $\tilde{c}_y)$ denotes the bounding box center coordinate relative to the input point cloud mean, $w$ and $l$ are respectively the width and length of the box and $[\cos(2\theta), \sin(2\theta)]^T$ represents the orientation angle vector, from which $\theta$ can be recovered. The estimation of these variables will be discussed as following:


\subsubsection{Angle Estimation}
\label{sec:angle}
Previous learning-based 3D bounding box estimation methods in the literature estimate the heading angle in one of the following ways: methods such as in \cite{ChenX18, ChenX16} propose a deep network to directly predict the heading angle from the input features. These methods should deal with the problem of angle discontinuity at $\pm \pi$. In the case of a horizontal bounding box, the network cannot learn if it should output $+\pi$ or $-\pi$ as a value of the estimated angle. To bypass this ambiguity, the second type of methods \cite{mousavian, charles1} use a classification-regression strategy, in which a classification head of the network estimates the angle bin index, while a regression head estimates the angle residue (difference) from the respective bin. The third type of methods, such as in \cite{ku},  predict the unit vector $[\cos(\theta), \sin(\theta)]^T$ as the estimate of the angle in the range $[-\pi,+\pi]$. This can speed up the box estimation procedure without sacrificing the accuracy.

Although angle estimation of 2D bounding box in BEV is similar to that in the above methods, the major difference is that in 2D bounding box fitting, the heading does not need to be estimated. Unlike the method proposed in \cite{ku} that fuses the dense ordered RGB information with LiDAR, there is not enough information for the network to accurately learn the heading in bounding box fitting with sparse 2D BEV LiDAR points. As a result, by ignoring the heading information, angle $\theta$ should only be estimated in the range $(-\frac{\pi}{2},\frac{\pi}{2}]$ or $[0,\pi)$ as in \cite{kim}. Throughout the paper, we may use ``orientation" and ``angle" interchangeably in the statements.

In the proposed BoxNet method, $\theta\in (-\frac{\pi}{2},\frac{\pi}{2}]$ is estimated using a regression head only. Due to angle discontinuity at $\theta=\pm \frac{\pi}{2}$, we train the network to learn $\sin(2\theta)$, which is continuous at $\theta=\pm \frac{\pi}{2}$. Although using this function solves the discontinuity problem, it is not sufficient to uniquely estimate $\theta$. Clearly, both $\theta=\frac{\pi}{4}-\epsilon$ and $\theta=\frac{\pi}{4}+\epsilon$ yield the same value for $\sin(2\theta)$, if $0<\epsilon<\frac{\pi}{4}$. To obtain a unique value for $\theta$,  the network is trained to learn both $\sin(2\theta)$ and $\cos(2\theta)$. As shown in Fig.~\ref{fig:boxnet}, to guarantee that the network output for $\sin(2\theta)$ and $\cos(2\theta)$ is in the range $[-1,1]$, ``$\tanh$'' is used as an activation function for the angle estimation head (branch). At inference time, angle is estimated using $\theta=\arctantwo (\sin(2\theta),\cos(2\theta))/2$.

As shown in Section \ref{abla}, using a double-angle sinusoidal head $[\cos(2\theta), \sin(2\theta)]^T$ yields a more accurate estimate of the bounding box orientation $\theta$ than a single-angle sinusoidal head $[\cos(\theta), \sin(\theta)]^T$.

\subsubsection{Size Estimation}
One of the main advantages of learning-based bounding box fitting methods over classical methods is their capability to learn the object geometry, or size from the training data. The proposed BoxNet learns the width $w$ and length $l$ of a bounding box using a size estimation head, as shown in  Fig.~\ref{fig:boxnet}. To guarantee that the network output for size is always positive, ``ReLU" is used as the activation function in the size estimation head.

\subsubsection{Center Estimation}
\label{concat}
Instead of directly estimating the object center, its center relative to the point cloud mean (or median) is predicted. As shown in Fig.~\ref{fig:boxnet},  point cloud mean that is estimated by mean pool operation, is subtracted from each point. Feature vector is then extracted from mean-reduced points using a feature extraction network consisting of several {\it multi-layer perceptron (MLP)} layers. Since the estimation of center is highly correlated with that of angle and size, we concatenate the estimated angle and size values with the extracted feature vector. The concatenated feature vector is then fed into a box center regressor, in which relative bounding box center $(\tilde{c}_x$, $\tilde{c}_y)$ can be estimated. Absolute box center $(c_x, c_y)$ is then obtained by adding the point cloud mean (or median) to $(\tilde{c}_x$, $\tilde{c}_y)$.

\subsection{Loss Function}
The regression loss for the proposed BoxNet is obtained as a weighted average loss of $\theta$, $\{w,l\}$ and $\{c_x, c_y\}$, given by
\begin{equation}
\label{eq:loss}
loss = w_0\cdot L(\theta)+w_1\cdot L(\{w,l\})+w_2\cdot L(\{c_x, c_y\})
\end{equation}
where $L(.)$ denotes a regression loss function, such as {\it mean squared error} (MSE) or Huber loss. 
The optimum weight values, obtained via cross-validation, are $w_0=w_2=1$ and $w_1=2$.

\section{Experiments}
\label{experiment}

This section discusses the experimental results of the proposed method, validated on the KITTI 3D benchmark \cite{geiger}, which provides high-quality object annotation and sensor calibration.

\subsection{Data Preprocessing}

The KITTI 3D object dataset has 7481 image frames and their corresponding LiDAR point clouds, each of which contains one or multiple objects. For each LiDAR frame, we first transformed the point cloud to the camera coordinate system and then extracted all the points within every ground truth bounding box according to the annotation of that frame. As in most autonomous driving research, only 3 classes of objects, i.e. {\it car}, {\it pedestrian} and {\it cyclist} were evaluated in our study. We excluded cases with few points as they do not provide sufficient information in recognizing an object, and only retained objects with more than 30 points. Vertical coordinate was then discarded to produce the BEV points and their 2D bounding boxes. Table \ref{dataset} shows the data splits used to train and test the proposed BoxNet.

\begin{table}[h]
\caption{Number of training and testing samples}
\label{dataset}
\begin{center}
\begin{tabular}{c|c|c|c|}
\cline{2-4}
& Car & Pedestrian & Cyclist \\
\hline
\multicolumn{1}{ |c|  }{Training} & 15,000 & 2700 & 825 \\
\hline
\multicolumn{1}{ |c|  }{Testing} & 5000 & 931 & 276 \\
\hline
\multicolumn{1}{ |c|  }{Total} & 20,000 & 3631 & 1101 \\
\hline
\end{tabular}
\end{center}
\end{table}

As mentioned in section~\ref{sec:angle}, by removing the heading information, object angles are normalized to be in the range $(-\frac{\pi}{2},\frac{\pi}{2}]$. Since samples in the training and testing sets have different numbers of points, multiple samples cannot be processed in one batch. To enable batch processing of the data, each object point cloud is resampled to $N=512$ points.

\subsection{Training the Neural Network}
Fig.~\ref{fig:boxnet} shows the details of the BoxNet structure. The feature extraction network is composed of shared MLP (64, 128, 1024) and a max-pooling layer across points. The feature vector of size 1024 was fed into a box regression network which consists of 3 regression heads (for orientation, size and center regression). Each head consists of {\it fully connected layers} (FCL) of size (512, 128, 2). All the fully connected layers include batch normalization. ``Tanh", ``ReLU" and ``linear" activation functions were used for the last FCL of orientation, size and center regression heads, respectively. We used the Adam optimizer with an initial learning rate 0.005 and exponentially decayed it with a rate 0.7 and step 250,000.  The decay rate of batch normalization started from 0.5 and was increased to 0.99. For 15,000 car samples, training the network with TensorFlow on a P100 GPU with batch size of 32, took 2-3 hours for 400 epochs.

\subsection{Evaluation Criteria}
\label{metrics}
We are interested in understanding how well an estimated bounding box can align with its ground truth, in terms of both pose and size, and used the following metrics to evaluate the results:
\begin{itemize}
\item \textbf{Center error}: The absolute difference between box centers,
\begin{equation}
err_c = \sqrt{(x_p-x_g)^2+(y_p-y_g)^2}
\end{equation}
where $(x_p,y_p)$ and $(x_g,y_g)$ are centers of the estimated and ground truth bounding boxes, respectively.
\item \textbf{Orientation error}: The smaller rotation amount from the estimated box to the ground truth one,
\begin{equation}
\label{theta_eval}
err_\theta = \begin{cases}
\theta_g-\theta_p, & \text{if $-\frac{\pi}{2} \leq \theta_g-\theta_p \leq \frac{\pi}{2}$}\\
\theta_g-\theta_p-\pi, & \text{if $\theta_g-\theta_p > \frac{\pi}{2}$}\\
\theta_g-\theta_p+\pi, & \text{if $\theta_g-\theta_p <-\frac{\pi}{2}$}\\
\end{cases}
\end{equation}
where $\theta_p$ and $\theta_g$ are the orientation angles. Eq. (\ref{theta_eval}) will ensure the angle error between $(-\frac{\pi}{2}, \frac{\pi}{2}]$. Note that in calculating the average error of all the testing samples, we used $|err_\theta|$ instead of $err_\theta$ and the results are reported in degrees instead of radians.
\item \textbf{Intersection over Union (IoU)}: The overlap between two boxes in terms of area, relative to the combination,
\begin{equation}
IoU = \frac{S_p \cap S_g}{S_p \cup S_g}
\end{equation}
where $S_p$ and $S_g$ represent the area. Here we do not directly compare the width and length because it is the normal case for the method in \cite{zhang} to mix up the two edges due to its inability to predict the size, resulting in a $\pi/2$ bias in estimating the angle. Therefore, the IoU would be a better alternative to evaluate the size.
\end{itemize}

The above criteria are widely used in computer vision and robotics research, to evaluate the algorithm performance in applications such as object detection and robot localization. A smaller error and larger IoU indicate better performance of the method.

\subsection{Statistical Results}

\begin{figure*}
\centering
\begin{subfigure}{1.0\columnwidth}
\includegraphics[width=\columnwidth]{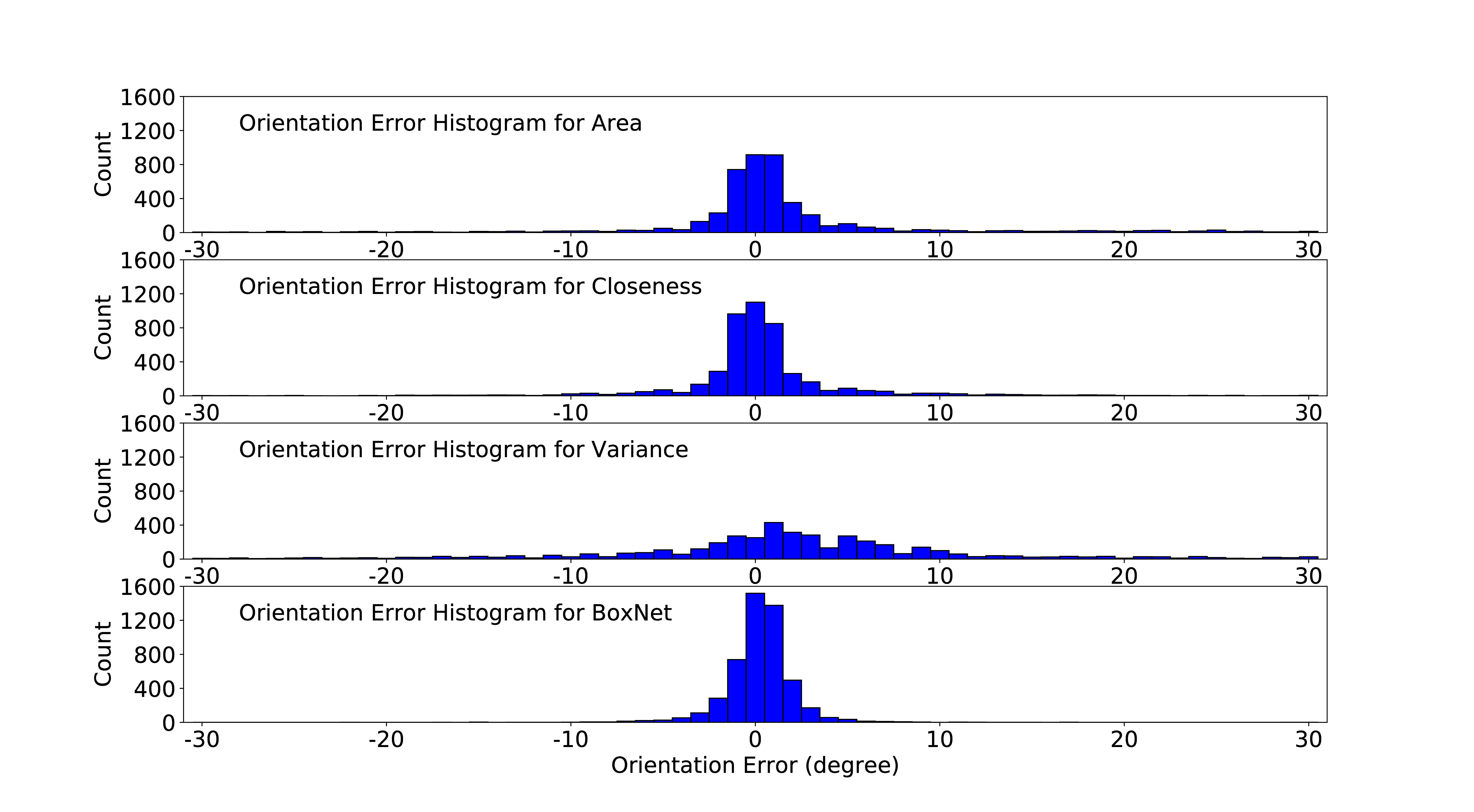}
\label{car_err}
\end{subfigure}
\begin{subfigure}{1.0\columnwidth}
\includegraphics[width=\columnwidth]{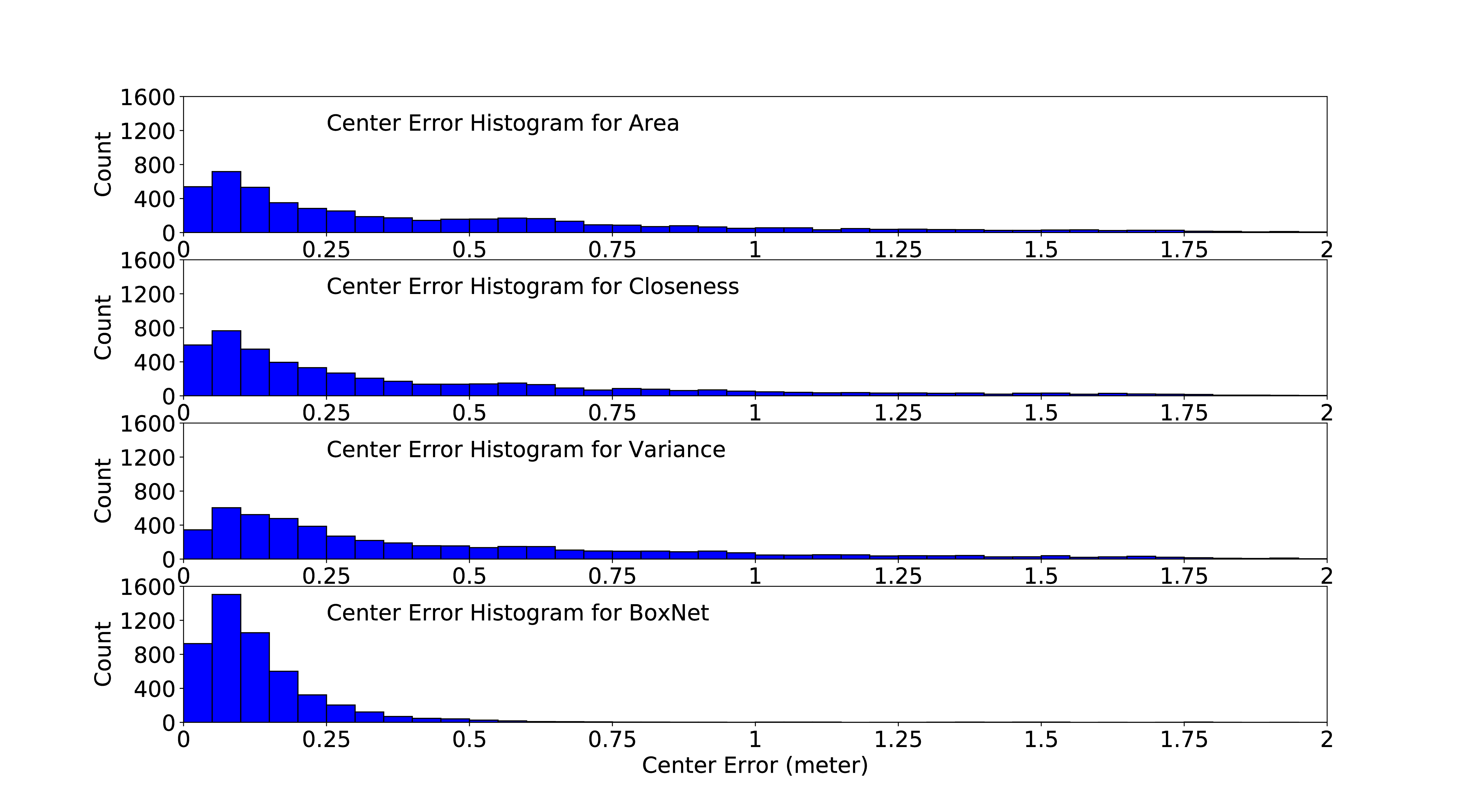}
\label{cyc_err}
\end{subfigure}
\caption{Distribution of orientation error (left) and center error (right) for the car category. BoxNet has significantly improved performance compared with the SLF method in three versions. See Table \ref{error_center_theta} and Table \ref{iou} for a detailed comparison in terms of the average error and IoU for all three classes.}
\label{error_cen_ori}
\end{figure*}

The proposed BoxNet method was thoroughly compared with the state-of-the-art 2D bounding box fitting algorithm {\it Search-based L-shape Fitting (SLF)}  \cite{zhang}, in which {\it area}, {\it closeness} and {\it variance} were used as the objectives to optimize the box that can tightly embrace all the points -- a non-learning generic strategy focusing on the observed points without considering the correlation prior between points and object geometry. In order to have a fair comparison, we implemented and ran the SLF method only on the testing samples. Fig. \ref{error_cen_ori} shows the distribution (in histogram) of both orientation (in degrees) and center (in meters) errors of all the testing samples for the car category. BoxNet has obviously better performance in estimating the orientation compared with different versions of SLF. The latter, especially when using variance, has a much wider distribution of orientation error. Similar performance boost can be observed in the estimation of box center location, where BoxNet significantly improves the accuracy.

\begin{table}[h]
\caption{Average error in center (in meters) and orientation (in degrees) of three classes. Comparison results between SLF \cite{zhang} (area, closeness and variance) and BoxNet.}
\label{error_center_theta}
\begin{center}
\begin{tabular}{cc|c|c|c|c|c}
\cline{3-6}
& & Area & Closeness & Variance & BoxNet \\
\cline{1-6}
\multicolumn{1}{ |c  }{\multirow{2}{*}{Car} } &
\multicolumn{1}{ |c| }{$\overline{err_c}$} & 0.4305 & 0.3941 & 0.4494 & \textbf{0.1401} &     \\
\cline{2-6}
\multicolumn{1}{ |c  }{}                        &
\multicolumn{1}{ |c| }{$\overline{|err_\theta|}$} & 9.6027 & 8.4223 & 14.5245 & \textbf{1.8057} &     \\
\cline{1-6}
\multicolumn{1}{ |c  }{\multirow{2}{*}{Cyclist} } &
\multicolumn{1}{ |c| }{$\overline{err_c}$} &\textbf{0.1020} & 0.1046 & 0.1042 & 0.1046 &  \\
\cline{2-6}
\multicolumn{1}{ |c  }{}                        &
\multicolumn{1}{ |c| }{$\overline{|err_\theta|}$} & 6.9517 & 7.1199 & 9.0141 & \textbf{2.7773} &  \\
\cline{1-6}
\multicolumn{1}{ |c  }{\multirow{2}{*}{Pedestrian} } &
\multicolumn{1}{ |c| }{$\overline{err_c}$} & 0.0814 & 0.0822 & \textbf{0.0812} & 0.1031 &  \\
\cline{2-6}
\multicolumn{1}{ |c  }{}                        &
\multicolumn{1}{ |c| }{$\overline{|err_\theta|}$} & 43.4191 & 42.2572 & 43.9135 & \textbf{18.6729} &  \\
\cline{1-6}
\end{tabular}
\end{center}
\end{table}

\begin{figure*}
\centering
\begin{subfigure}{1\columnwidth}
\includegraphics[width=\columnwidth]{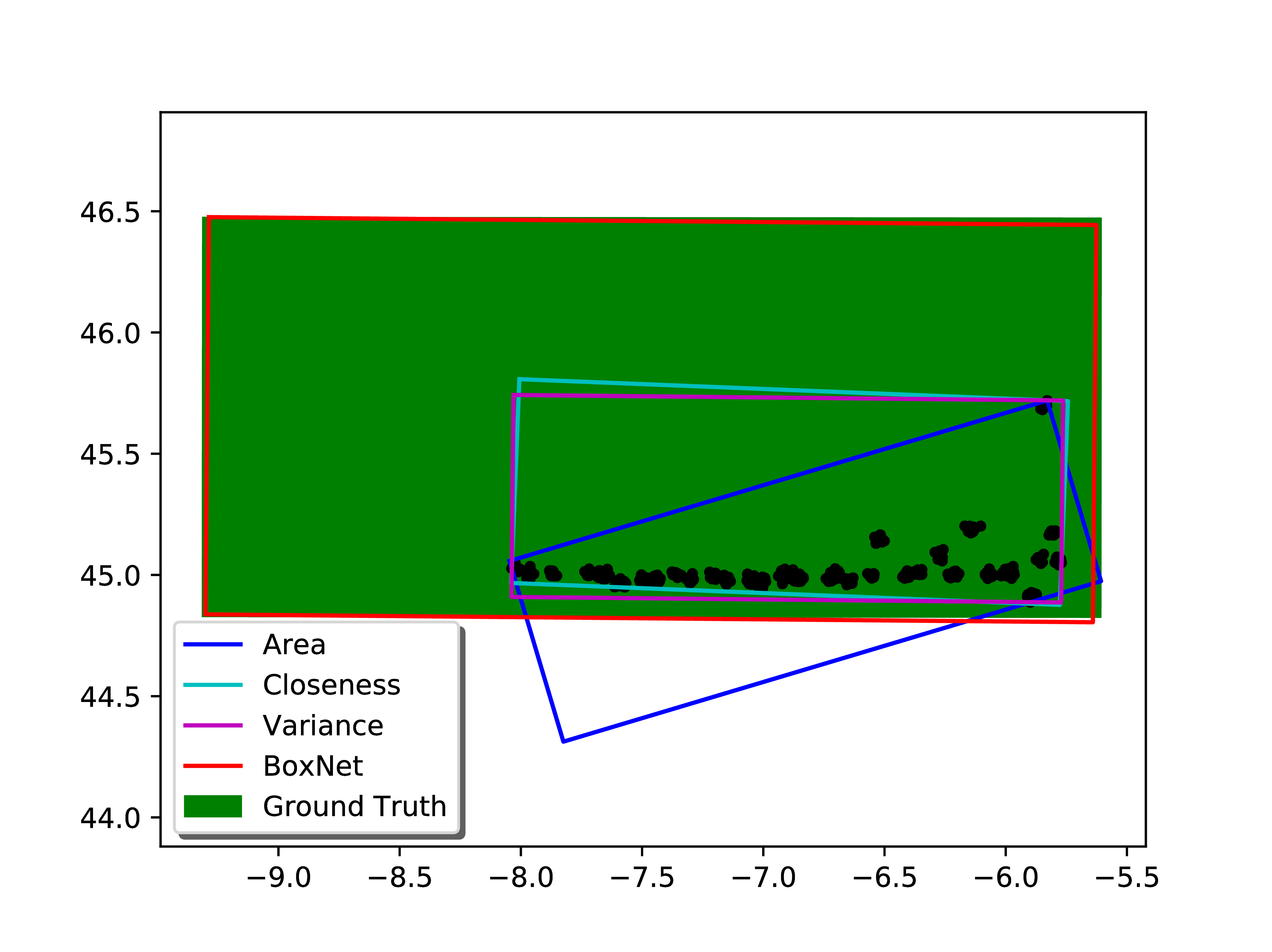}
\caption{}
\label{car1}
\end{subfigure}
\begin{subfigure}{1\columnwidth}
\includegraphics[width=\columnwidth]{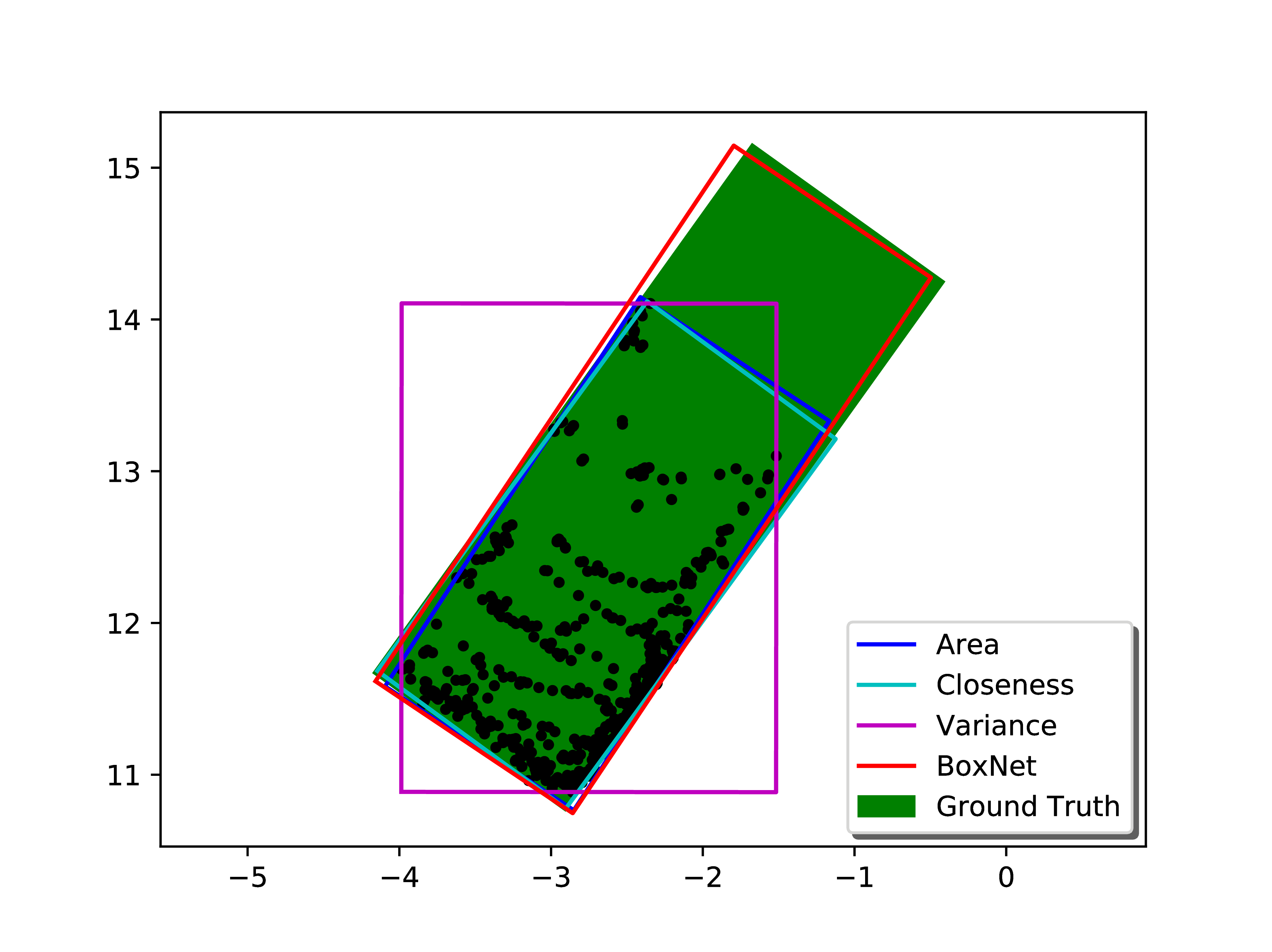}
\caption{}
\label{car2}
\end{subfigure}
\begin{subfigure}{1\columnwidth}
\includegraphics[width=\columnwidth]{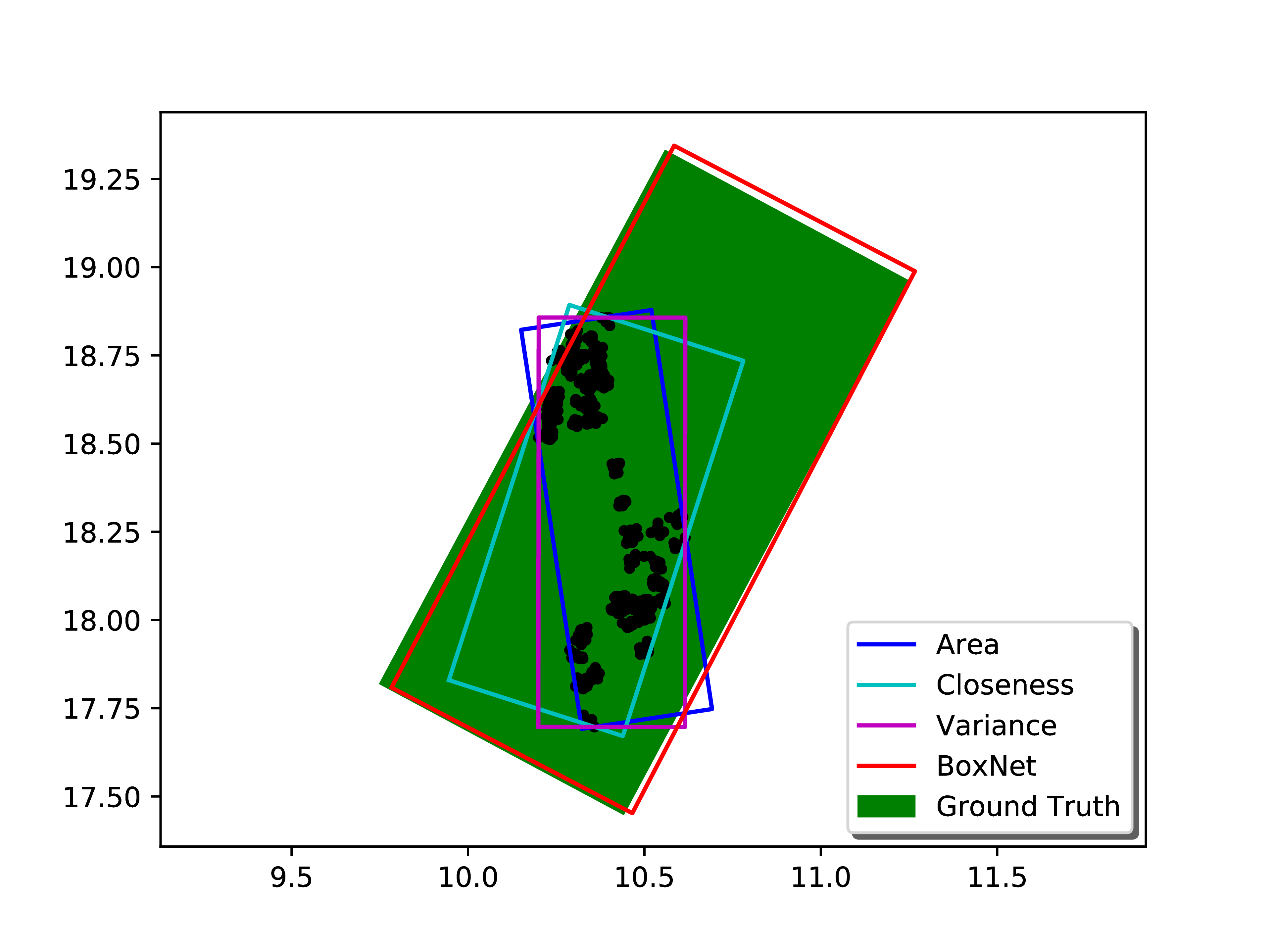}
\caption{}
\label{cyclist1}
\end{subfigure}
\begin{subfigure}{1\columnwidth}
\includegraphics[width=\columnwidth]{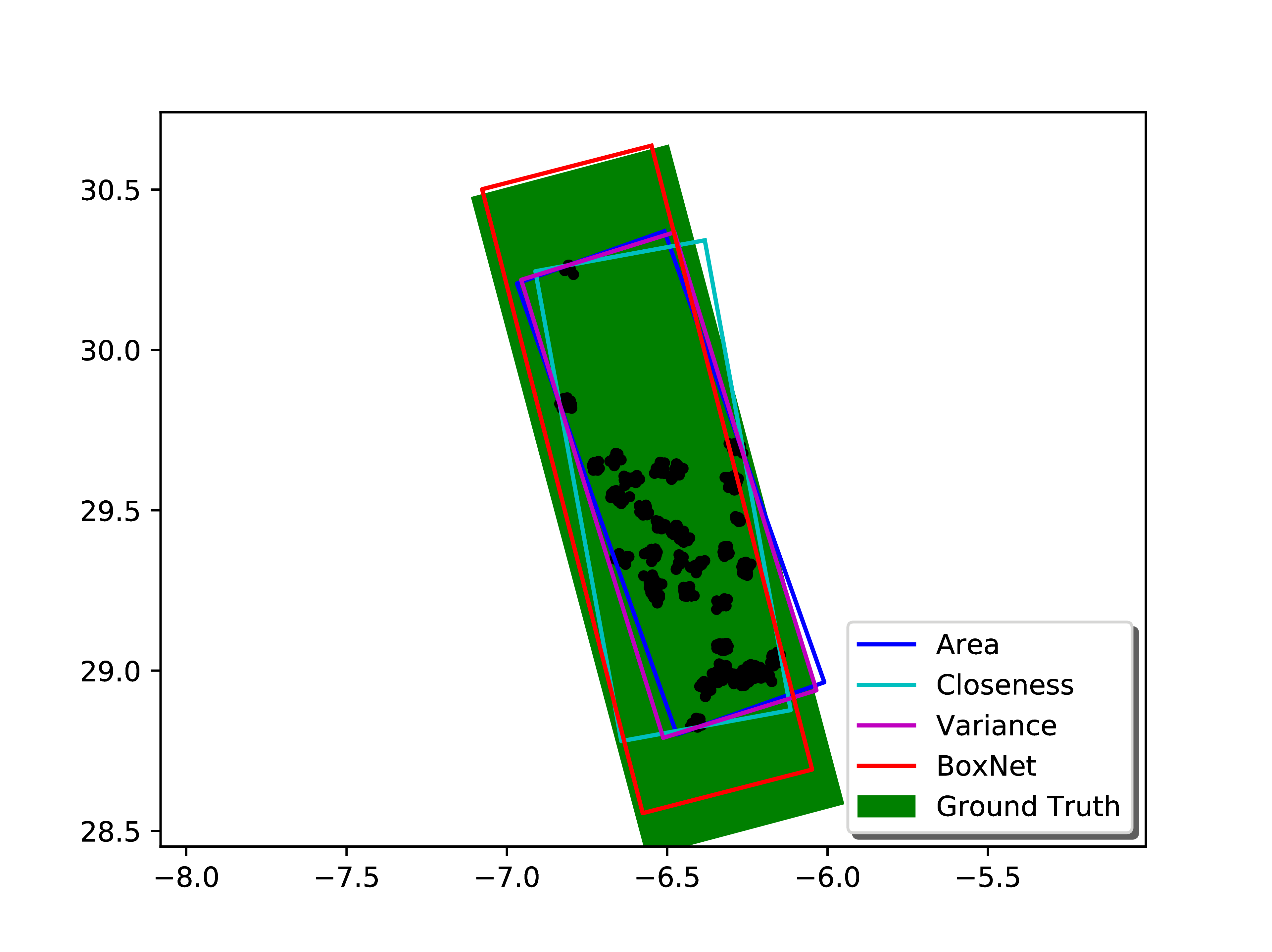}
\caption{}
\label{cyclist2}
\end{subfigure}
\begin{subfigure}{1\columnwidth}
\includegraphics[width=\columnwidth]{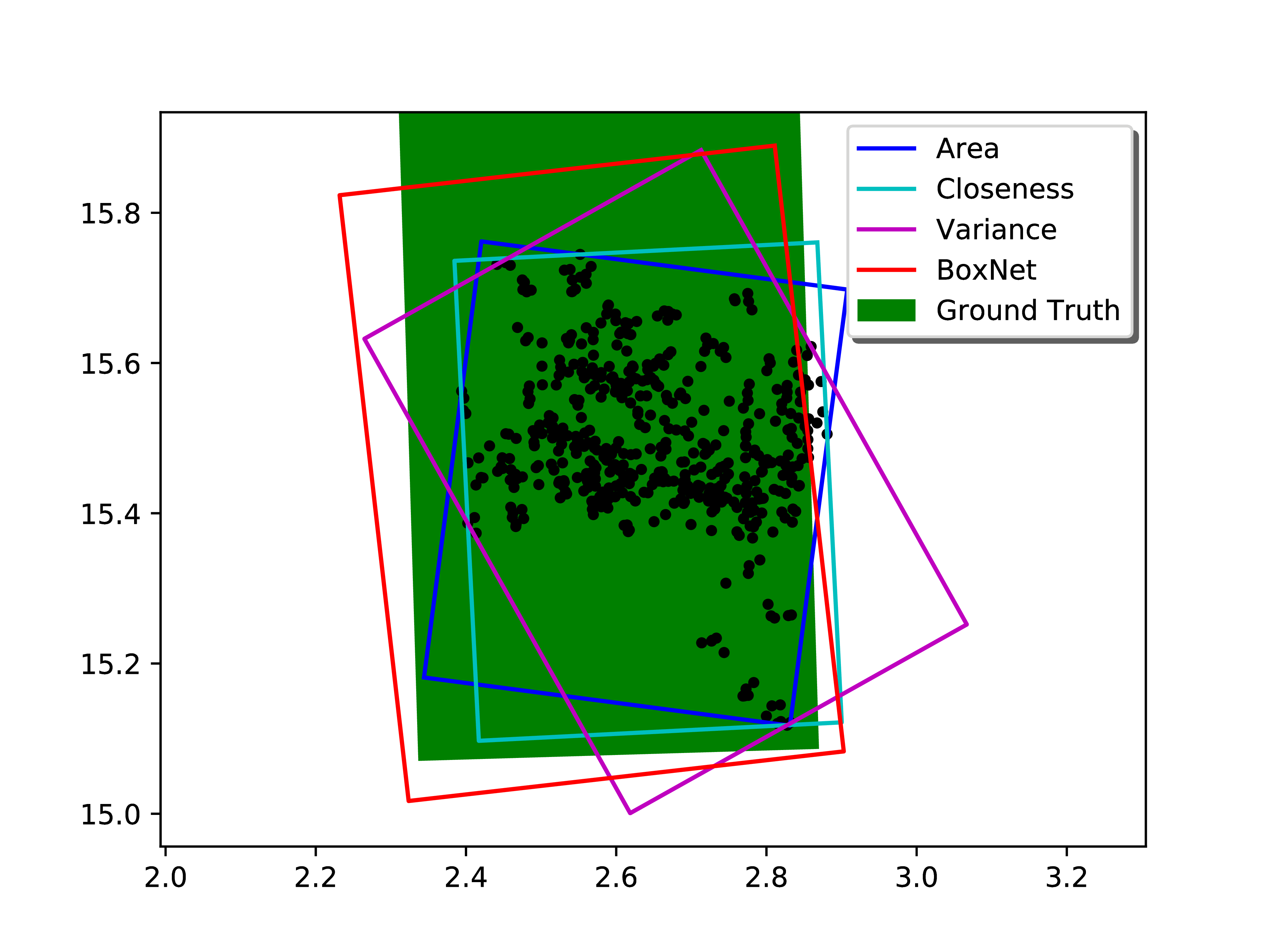}
\caption{}
\label{ped1}
\end{subfigure}
\begin{subfigure}{1\columnwidth}
\includegraphics[width=\columnwidth]{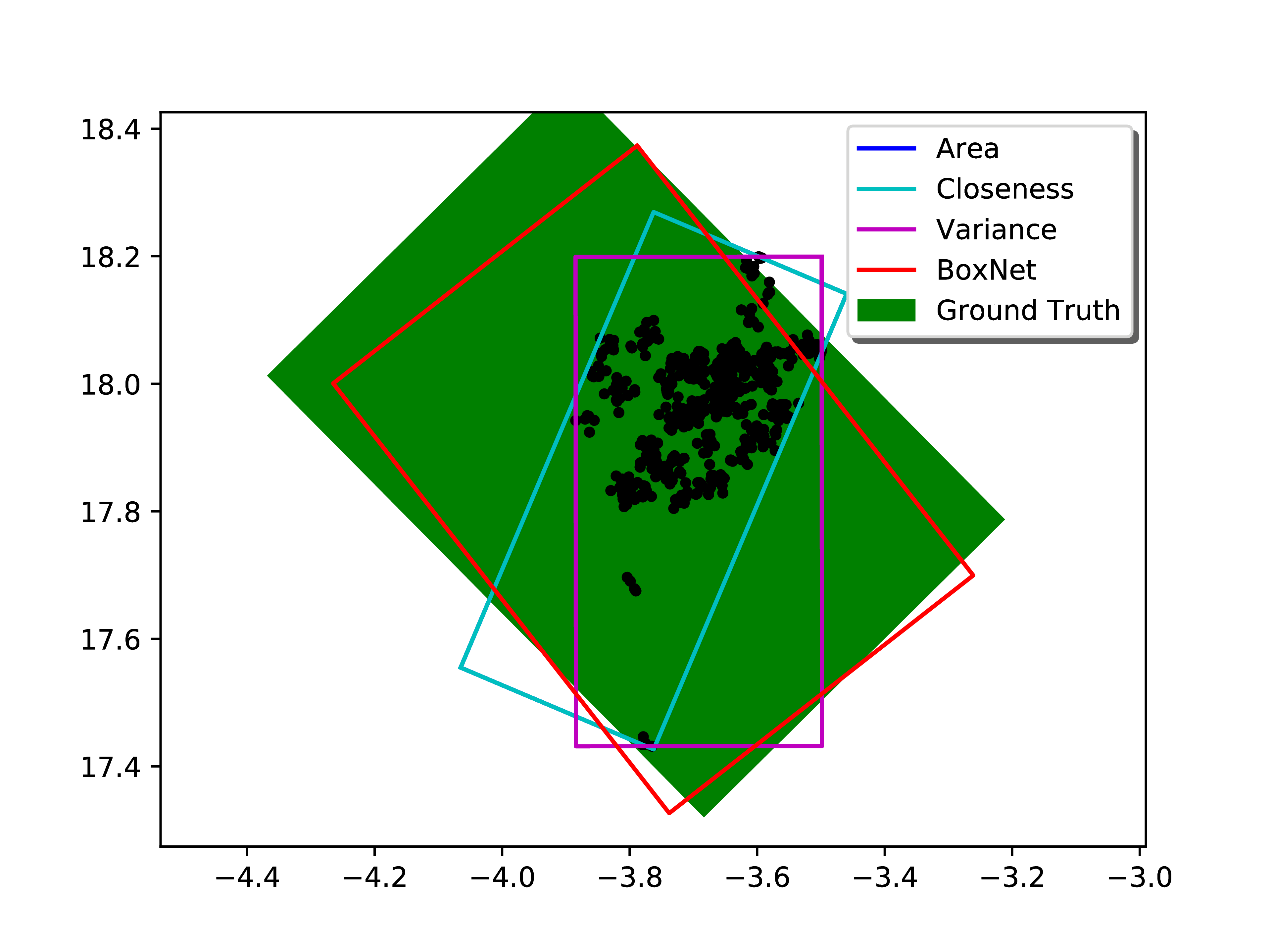}
\caption{}
\label{ped2}
\end{subfigure}
\caption{Qualitative examples in bounding box estimation (best in color). (a) and (b): Car, (c) and (d): Cyclist, (e) and (f): Pedestrian. Bounding boxes are labeled in different colors and ground truth is shown in a filled rectangle. The state-of-the-art SLF method \cite{zhang} in its different versions of {\it area}, {\it closeness} and {\it variance} cannot predict the actual object geometry with insufficient 2D points. Moreover, the criteria may not be consistently reliable in different scenarios. The proposed BoxNet has better performance in estimating the box location and geometry. Note that it is generally more difficult to estimate the bounding box of a pedestrian than a car, yet our method still works fairly well for the category of pedestrian. In (f), area is invisible due to its overlap with closeness since both of them give the same estimation results.}
\label{examples}
\end{figure*}

The advantage of BoxNet is also reflected in the other two classes, as summarized in Table \ref{error_center_theta}, where average errors in orientation and center estimation are shown. While the improvement is obvious in estimating the orientation in all the classes, we have observed that the average error in the pedestrian category is still one order of magnitude larger than that in the other classes. This is understandable because the bounding box of a pedestrian is much smaller and the length-to-width ratio is closer to 1. As a result, BEV points do not indicate a clear primary axis along which person's facing direction is defined without ambiguity. Therefore, it becomes more difficult to estimate the orientation of a person. On the other hand, the other two classes have a larger length-to-width ratio, which helps determine the angle more accurately.

In terms of location, BoxNet reduces the average error from 45cm to 14cm for the car category. It provides comparable results with the SLF method in pedestrian and cyclist categories, with slightly lower performance of 2cm difference at most. However, in terms of IoU summarized in Table \ref{iou}, BoxNet has a huge performance boost over the others, due to its superiority in estimating the object geometry. As will be shown in the qualitative examples in Section \ref{qualitative}, BoxNet can predict the actual size of the object with accurate width and length values, while the SLF method fails. The reason why SLF is better than BoxNet in estimating the center of a pedestrian may also be due to the small size of the box that favors the former method's calculation of the box center.

\begin{table}[h]
\caption{Average IoU of three classes. Comparison results between SLF \cite{zhang} (area, closeness and variance) and BoxNet.}
\label{iou}
\begin{center}
\begin{tabular}{c|c|c|c|c|}
\cline{2-5}
& Area & Closeness & Variance & BoxNet \\
\hline 
\multicolumn{1}{ |c|  }{Car} & 0.6578 & 0.6825 & 0.6346 & \textbf{0.8787}\\
\hline
\multicolumn{1}{ |c|  }{Cyclist} & 0.622 & 0.6252 & 0.622 & \textbf{0.7953} \\
\hline
\multicolumn{1}{ |c|  }{Pedestrian} & 0.5209 & 0.5368 & 0.549 & \textbf{0.6704} \\
\hline
\end{tabular}
\end{center}
\end{table}

\subsection{Qualitative Results}
\label{qualitative}

Figure \ref{examples} shows qualitative examples of bounding box estimation using the proposed BoxNet and the state-of-the-art SLF method for the three classes of {\it car}, {\it cyclist} and {\it pedestrian}. When a small set of points are observable, BoxNet has much better performance in predicting the actual size of the object due to the prior knowledge it has obtained from training samples, while SLF will always fit a tight box to the given 2D points, without extending the width or length to match the object's real size. In addition, the result of using area, closeness or variance in the SLF method may vary significantly in different scenarios, making it difficult to choose a criterion that can perform well for all the cases. Our method, on the other hand, is generally robust and reliable, and it is insensitive to the distribution of 2D points.

\subsection{Ablation Studies}
\label{abla}

The ablation studies in this section focus on demonstrating the effectiveness of the proposed solutions to the orientation discontinuity and box location estimation. In these studies, we only ran the experiments for the car category. The results can be generalized to the other two classes.

\textbf{Orientation Improvement} To tackle the orientation discontinuity problem, we have proposed to estimate the double-angle sinusoidal vector $[\cos(2\theta), \sin(2\theta)]^T$, instead of directly estimating the angle itself. To demonstrate the effectiveness of this technique, we compare $err_d$, $|err_\theta|$ and IoU, all in average values between the estimated and ground truth bounding boxes for the following cases: $\theta$, $[\cos(\theta), \sin(\theta)]^T$ and $[\cos(2\theta), \sin(2\theta)]^T$. Results are summarized in Table \ref{discontinuity}, where point cloud mean is subtracted  from the 2D points for all the samples. Clearly, $[\cos(2\theta),\sin(2\theta)]^T$ solution yields the smallest orientation error and highest IoU, showing its superiority to the other two solutions. An interesting phenomenon is that by estimating $\theta$ directly, the average orientation error is larger than that of the SLF method, yet the IoU is higher. This is due to the accurate estimation of the size and center location in BoxNet, as described below.

\begin{table}[h]
\caption{Comparison of methods in resolving orientation discontinuity (average error and IoU for the car category)}
\label{discontinuity}
\begin{center}
\begin{tabular}{|c|c|c|c|c|}
\hline
Method & Evaluation Loss & $\overline{err_c}$ & $\overline{|err_\theta |}$ & $\overline {IoU}$\\
\hline\hline
$\theta$ & 0.1606 & 0.1427 & 15.0999 & 0.7311\\
\hline
$\cos(\theta),\sin(\theta)$ & 0.1060 & 0.1369 & 3.0812 & 0.8539\\
\hline
$\cos(2\theta),\sin(2\theta)$ & 0.0723 & 0.1401 & 1.8057 & \textbf{0.8787}\\
\hline
\end{tabular}
\end{center}
\end{table}

\textbf{Center Improvement} Bounding box center location is estimated by subtracting the point cloud mean/median from the original 2D point coordinates. To show the effectiveness of this solution, we compare the results with that of no mean/median subtraction. The results in terms of average error and IoU are shown in Table \ref{center_est}. In this experiment, $[\cos(2\theta), \sin(2\theta)]^T$ was used to estimate the orientation. While subtracting mean/median provides better performance in estimating the orientation ($2.5^{\circ}$ improvement), it also significantly improves the accuracy of center estimation by more than 25cm. On the other hand, we can see that mean and median generate comparable results. While mean value is more efficient to compute (no need to sort the points according to their coordinates), median may be a better choice when data is contaminated by outliers of long-tailed nature, such as salt-and-pepper noise.

\begin{table}[!t]
\caption{Comparison of center estimation methods (average error and IoU for the car category)}
\label{center_est}
\begin{center}
\begin{tabular}{|c|c|c|cc|}
\hline
Method & Evaluation Loss & $\overline {err_d}$ & $\overline{|err_\theta|}$ &$\overline{IoU}$\\
\hline\hline
None & 0.1478 & 0.3937 & 4.0252 & 0.7265\\
\hline
Median & 0.0722 & 0.1226 & 1.5291 & \textbf{0.8776}\\
\hline
Mean & 0.0723 & 0.1401 & 1.8057 & \textbf{0.8787}\\
\hline
\end{tabular}
\end{center}
\end{table}

\textbf{Feature Concatenation} The effect of feature concatenation discussed in Section \ref{concat} is shown in Table \ref{feature_concatenation}. The intuition behind this concatenation is that an accurate estimate of the box orientation and size can lead to a better estimate of the box center, due to the existing dependency between these variables. Effectiveness of the proposed technique is confirmed by the results shown in Table \ref{feature_concatenation}, where our solution improves the average IoU by 1.5\%.

\begin{table}[h]
\caption{The effect of feature concatenation}
\label{feature_concatenation}
\begin{center}
\begin{tabular}{|c|c|c|}
\hline
Method & Evaluation Loss & $\overline{IoU}$\\
\hline \hline
Without concatenation & 0.0751 & 0.8634 \\
\hline
With concatenation & 0.0723 & \textbf{0.8787}\\
\hline
\end{tabular}
\end{center}
\end{table}

\begin{figure}[!t]
     \centering
      \includegraphics[scale=0.28]{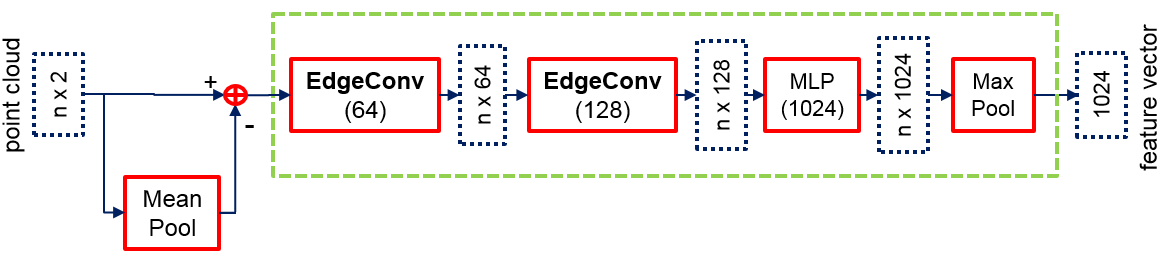}
      \caption{\textbf {EdgeConv-based BoxNet}. The first two MLP layers of the feature extraction network in baseline BoxNet were replaced with EdgeConv~\cite{edgeconv}.}
      \label{fig:boxneteconv}
\end{figure}

\begin{figure}[!t]
     \centering
      \includegraphics[scale=0.06]{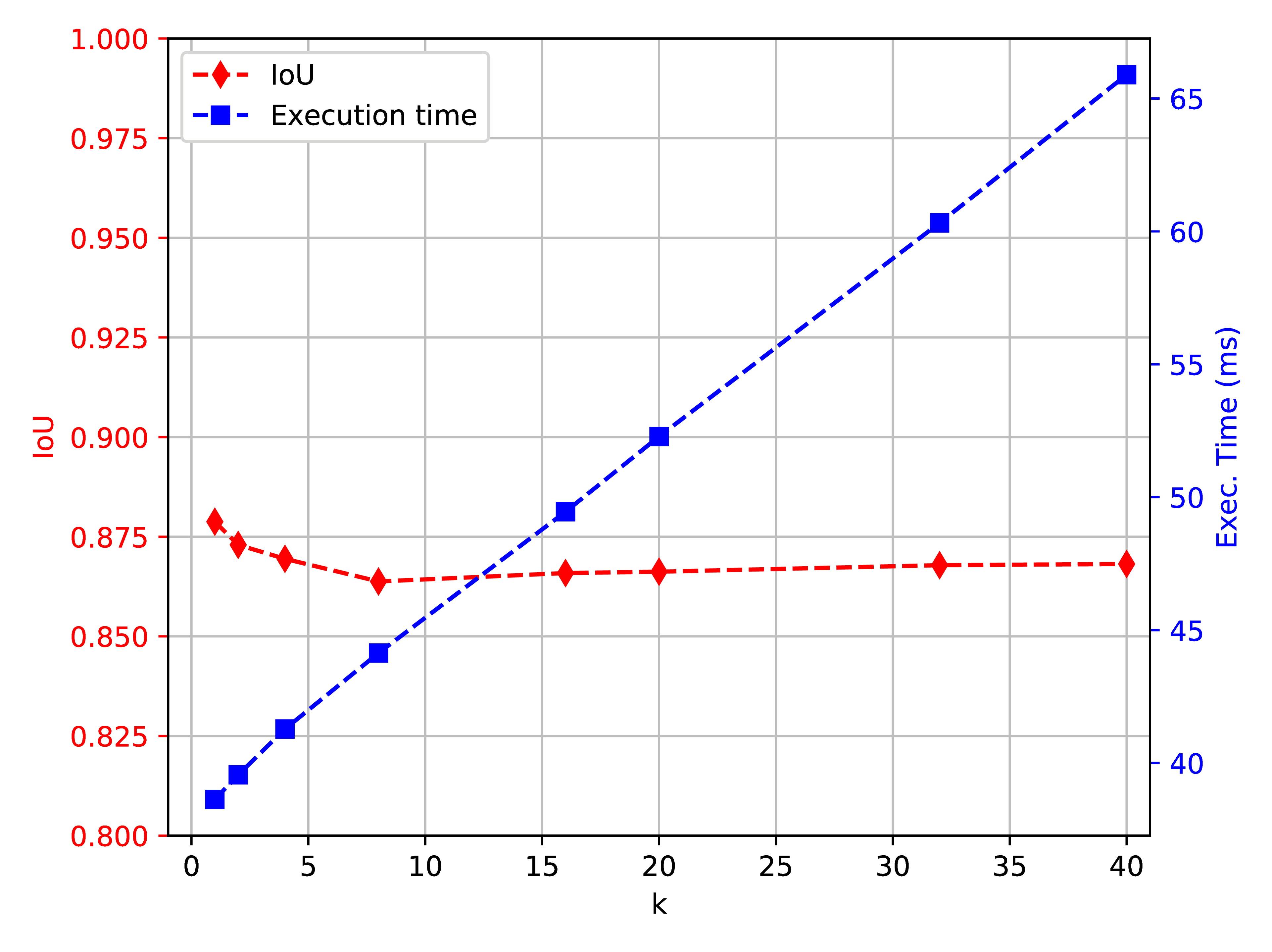}
      \caption{Effect of using $k$ nearest points in EdgeConv-based BoxNet on average IoU and execution time (in ms), for the Car category. All the experiments were conducted on a single P100 GPU with batch size of 32.}
      \label{fig:boxnet_econv_k}
\end{figure}

\textbf{Neighboring Points Information} To study the effect of using the information from neighboring points on the estimation accuracy, we replaced the first two MLP layers of the feature extraction network in the baseline BoxNet with EdgeConv \cite{edgeconv} layers, as shown in Fig. \ref{fig:boxneteconv}. Unlike an MLP layer where input points are filtered independently, EdgeConv defines a convolution-like operation at each point using its {\it k} nearest neighbors. Fig. \ref{fig:boxnet_econv_k} shows how average IoU changes with respect to {\it k}. As seen, unlike 3D point cloud object classification problem where EdgeConv surpasses PointNet in classification accuracy \cite{edgeconv}, in our bounding box regression problem, using the nearest neighbor points yields lower IoU values than the baseline BoxNet. This means that in the bounding box regression problem, local correlation between points inside an object does not have as much information as the boundary (the outmost set of) points. Note that not only the EdgeConv-based BoxNet falls behind the baseline BoxNet in box estimation accuracy, its computational complexity grows much higher as {\it k} increases.

\begin{table}[!b]
\caption{Execution time in inference mode on P100 GPU (results for car category)}
\label{tab:comp_complexity}
\begin{center}
\begin{tabular}{|c|c|}
\hline
Method & Execution Time (ms) \\
\hline \hline
Baseline, mean pool, batch size=1 & 6.0 \\ \hline
Baseline, median pool, batch size=1 & 6.123 \\ \hline
Baseline, mean pool, batch=32 & 8.5 \\ \hline
Scale=1/2, batch size=32 & 7.6 \\ \hline
Scale=1/4, batch size=32 & 7.4 \\ \hline
Scale=1/8, batch size=32 & 7.36 \\ \hline
Scale=1/16, batch size=32 & 7.36 \\ 
\hline
\end{tabular}
\end{center}
\end{table}

\textbf{Network Size} The last ablation study we are interested in is the relationship between network size and its performance. In this study, the size of each network layer was shrinked by a certain scale, relative to the network shown in Fig.~\ref{fig:boxnet} as the baseline. If the size of any layer turns out to be less than (or equal to) 1, the layer is removed. Fig. \ref{netscale} shows the effect of shrinking the network size on the average IoU. As seen, even if the network size is reduced to 1/16 of its original size, its IoU only decreases by no more than 2\%. This means that the problem of estimating the 2D bounding box from 2D sparse points can also be solved with a much smaller deep network, at the expense of losing a small portion of accuracy. Note that even by shrinking our baseline network by scale 1/128, the average IoU for the car category (i.e. 0.7826) is still higher than that of the SLF method (i.e. 0.6825). On the other hand, reducing the network size implies faster processing time of LiDAR points.

\subsection{Computational Complexity}
\begin{figure}[!t]
     \centering
      \includegraphics[scale=0.06]{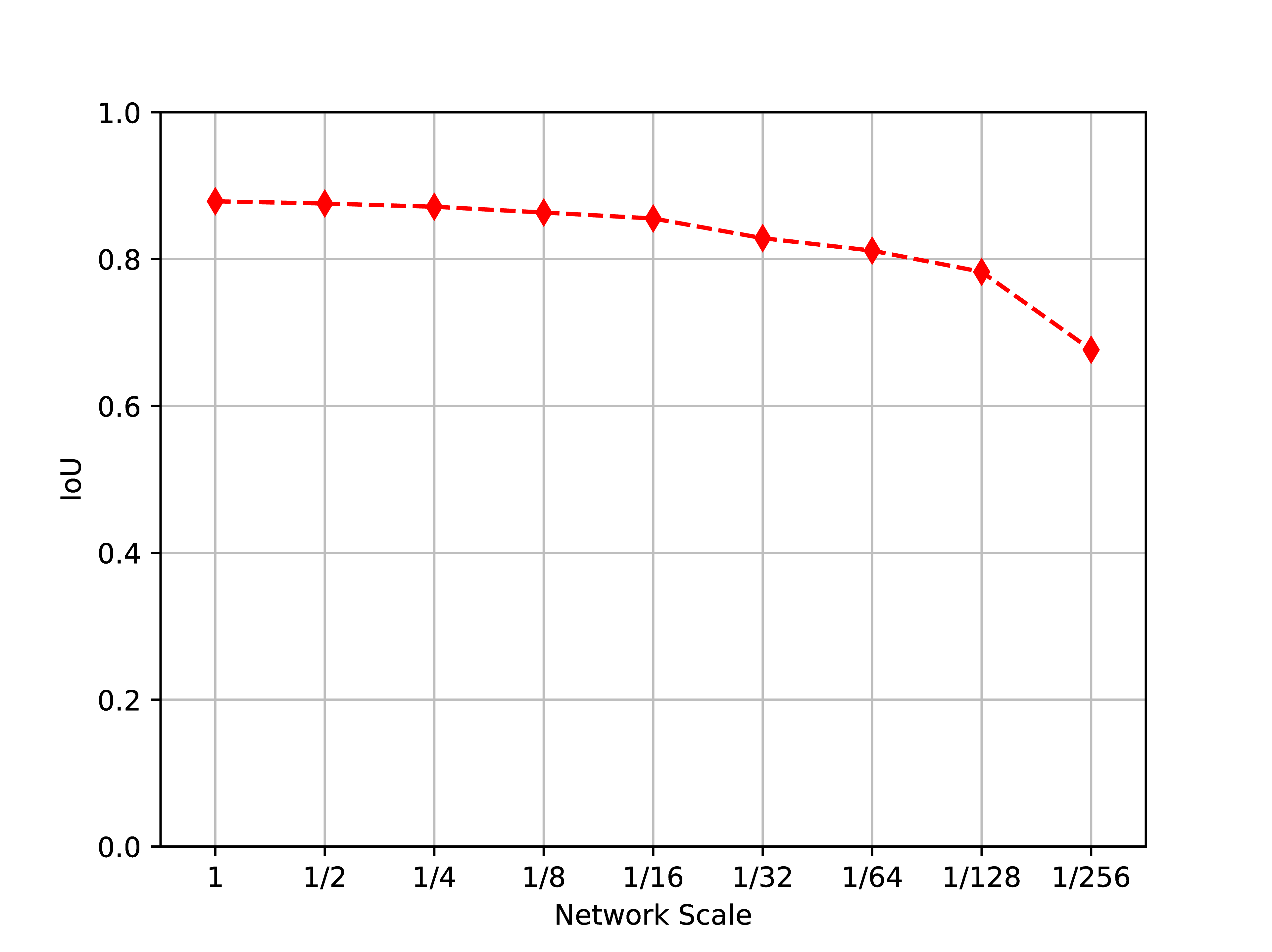}
      \caption{Effect of shrinking the network size on average IoU.}
      \label{netscale}
\end{figure}

Table~\ref{tab:comp_complexity} summarizes the execution time of running the experiments with different network structures, pooling methods and batch sizes on a single P100 GPU and a 1.2GHz CPU. As seen, using median takes slightly longer time to run than mean. The execution time of running the baseline BoxNet with batch size of 32 is 8.5ms, which is 42\% more than that with batch size of 1 point cloud (6ms), although the number of point clouds is 32 times more. This shows the effectiveness of using GPU in parallel processing and justifies our resampling of each point cloud to the fixed size of 512 points. Another interesting phenomenon shown in Table~\ref{tab:comp_complexity} is the execution time of running the shrinked BoxNet which is reduced to 7.6ms at scale 1/2, while saturating to 7.36ms at scales lower than 1/8. Based on the results given in Fig.~\ref{netscale} and Table~\ref{tab:comp_complexity}, we can conclude that BoxNet size can be shrinked by the scale of 1/16 without much reduction in IoU. This not only decreases the computational complexity but also highly reduces the model size.

\section{CONCLUSIONS}
\label{conclusion}

We have presented a method of estimating 2D bounding box from the bird's-eye view LiDAR points, or equivalently, 2D points from laser scanners. The proposed method utilizes a simple network to learn the correlation prior between object geometry and the points, and therefore is able to accurately predict the actual size of the object. To handle the orientation discontinuity that confuses the network in terms of learning the orientation, we used the sinusoid values of $[\cos(2\theta), \sin(2\theta)]^T$ to replace the original angle $\theta$ in the variables to be estimated. We also predicted the center relative to the point cloud mean to improve the estimation of box location. Experimental results show the superiority of the proposed method in estimating the bounding box orientation and size, compared with the state-of-the-art non-learning based method. It also performs well in predicting the box location and the IoU is significantly improved using the proposed method.

Future research will focus on extending the framework directly to 3D bounding box estimation from LiDAR points. We are also interested in exploring the possibility of using other network structures in bounding box regression.

\addtolength{\textheight}{-12cm}   









\end{document}